\documentclass[runningheads]{llncs}
\usepackage{graphicx}
\usepackage{comment}
\usepackage{color, colortbl}
\usepackage{amsmath,amssymb} %
\usepackage[table,dvipsnames]{xcolor}
\usepackage{wrapfig}
\usepackage{xspace}
\usepackage{multirow}
\usepackage{hyperref}
\usepackage[normalem]{ulem}

\usepackage[width=122mm,left=12mm,paperwidth=146mm,height=193mm,top=12mm,paperheight=217mm]{geometry}

\newcommand{\etal}{\textit{et al}.\xspace}
\newcommand{\dataset}{\mbox{\sc{SW}}i\mbox{\sc{G}}\xspace}
\newcommand{\imsitu}{\emph{imSitu}\xspace}
\newcommand{\task}{Grounded Situation Recognition\xspace}
\newcommand{\tasksmall}{\mbox{\sc{GSR}}\xspace}

\begin{document}
\pagestyle{headings}
\mainmatter

\title{Grounded Situation Recognition \vspace{-3mm}} %

\titlerunning{Grounded Situation Recognition}

\author{Sarah Pratt\inst{1} 
Mark Yatskar\inst{1} 
Luca Weihs\inst{1} 
Ali Farhadi\inst{2} 
Aniruddha Kembhavi \inst{1, 2}}
\authorrunning{S. Pratt et al.}

\institute{ \vspace{-3mm} \inst{1} Allen Institute for AI \quad \quad
\inst{2} University of Washington \\
\email{sarahp@allenai.org}
}

\maketitle
\vspace{-0.8em}

\vspace{-2mm}
\begin{abstract}
We introduce \task (\tasksmall), a task that requires producing structured semantic summaries of images describing: the primary activity, entities engaged in the activity with their roles (e.g. agent, tool), and bounding-box groundings of entities. \tasksmall presents important technical challenges: identifying semantic saliency, categorizing and localizing a large and diverse set of entities, overcoming semantic sparsity, and disambiguating roles. Moreover, unlike in captioning, \tasksmall is straightforward to evaluate. To study this new task we create the Situations With Groundings (\dataset{}) dataset which adds 278,336 bounding-box groundings to the 11,538 entity classes in the \imsitu dataset. We propose a Joint Situation Localizer and find that jointly predicting situations and groundings with end-to-end training handily outperforms independent training on the entire grounding metric suite with relative gains between 8\% and 32\%. Finally, we show initial findings on three exciting future directions enabled by our models: conditional querying, visual chaining, and grounded semantic aware image retrieval. Code and data available at \href{https://prior.allenai.org/projects/gsr}{ \color{blue} https://prior.allenai.org/projects/gsr}. \vspace{-1mm}

\vspace{-3mm}
\end{abstract}

\vspace{-5mm}
\begin{figure}
\centering

\includegraphics[width=12cm]{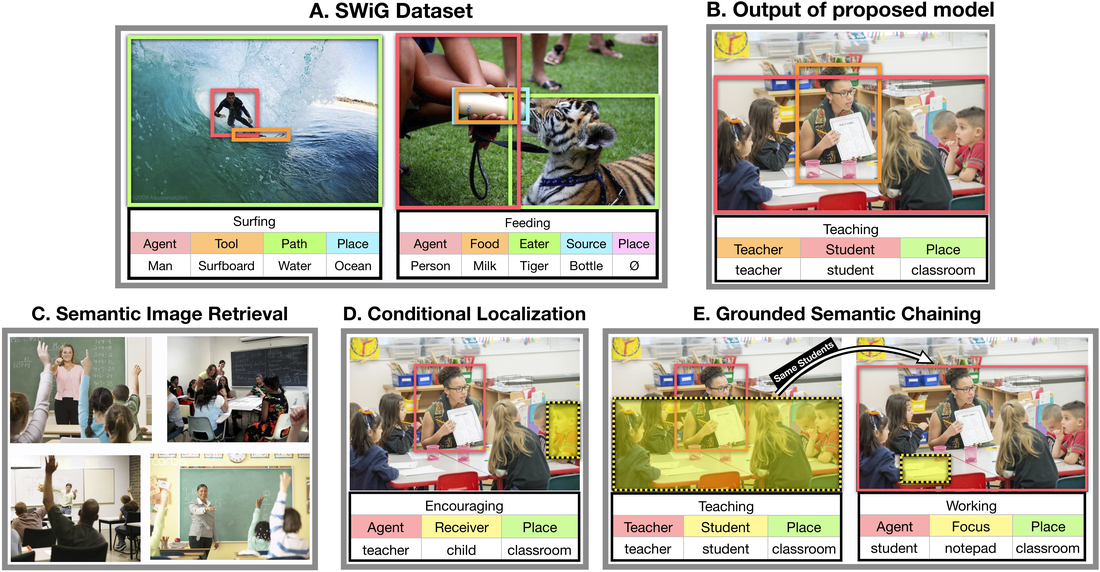}

\caption{\textbf{A} Two examples from our dataset: semantic frames describe primary activities and relevant entities. Groundings are bounding-boxes colored to match roles. \textbf{B} Output of our model (dev set image). \textbf{C} Top-4 nearest neighbors to \textit{B} using model predictions. Beyond visual similarity, these images are clearly semantically similar. \textbf{D} Output of the conditional model: given a bounding-box (yellow-dashed), predicts a relevant frame. \textbf{E} Example of grounded semantic chaining: given query boxes we are able to chain situations together. E.g. the teacher teaches students so they may work on a project
}

\label{fig:teaser}
\end{figure}

\section{Introduction}

Situation Recognition~\cite{imsitu} is the task of recognizing the activity happening in an image, the actors and objects involved in this activity, and the roles they play. The structured image descriptions produced by situation recognition are drawn from FrameNet~\cite{framenet}, a formal verb lexicon that pairs every verb with a frame of semantic roles, as shown in Figure~\ref{fig:teaser}. These semantic roles describe how objects in the image participate in the activity described by the verb. 

As such, situation recognition generalizes several computer vision tasks such as image classification, activity recognition, and human object interaction. It is related to the task of image captioning, which also typically describes the salient objects and activities in an image using natural language. However, in contrast to captioning, it has the advantages of always producing a structured and complete (with regards to semantic roles) output and it does not suffer from the well known challenges of evaluating natural language captions.

While situation recognition addresses \emph{what} is happening in an image, \emph{who} is playing a part in this and \emph{what} their roles are, it does not address a critical aspect of visual understanding: \emph{\textbf{where}} the involved entities lie in the image. We address this shortcoming and present \task (\tasksmall), a task that builds upon situation recognition and requires one to not just identify the situation observed in the image but also visually ground the identified roles within the corresponding image. \tasksmall presents the following technical challenges. \emph{Semantic saliency}: in contrast to recognizing all entities in the image, it requires identifying the key objects and actors in the context of the primary activity being presented. \emph{Semantic sparsity}: grounded situation recognition suffers from the problem of semantic sparsity~\cite{imsitucrf}, with many combinations of roles and groundings rarely seen in training. This challenge requires models to learn from limited data. \emph{Ambiguity}: grounding roles into images often requires disambiguating between multiple observed entities of the same category. \emph{Scale}: the scales of the grounded entities vary vastly with some entities also being absent in the image (in which case models are responsible for detecting this absence). \emph{Halucination}: labeling semantic roles and grounding them often requires halucinating the presence of objects since they may be fully occluded or off screen.

To train and benchmark models on \tasksmall, we present the \textbf{S}ituations \textbf{Wi}th \textbf{G}roundings dataset (\dataset) that builds upon the large \imsitu~dataset by adding 278,336 bounding-box-based visual groundings to the annotated frames. \dataset{} contains groundings for most of the more than 10k entity classes in \imsitu{} and exhibits a long tail distribution of grounded object classes. In addition to the aforementioned technical challenges of \tasksmall, the diversity of activities, images, and grounded classes, makes \dataset{} particularly challenging for existing approaches.

Training neural networks for grounded situation recognition using the challenging \dataset{} dataset requires localizing roughly 10k categories; a task that modern object detection models like RetinaNet~\cite{retinanet} struggle to scale to out of the box. We first propose modifications to RetinaNet that enables us to train large-class-cardinality object detectors. Using these modifications, we then create a strong baseline, the Independent Situation Localizer (ISL), that independently predicts the situation and groundings and uses late fusion to produce the desired outputs. Our proposed model, the Joint Situation Localizer (JSL), jointly predicts the situation and grounding conditioned on the context of the image. During training, JSL backpropagates gradients through the the entire network. JSL demonstrates the effectiveness of joint structured semantic prediction and grounding by improving both semantic role prediction and grounding and obtaining huge relative gains of between 8\% and 32\% points over ISL on the entire suite of grounding metrics.

Grounded situation recognition opens up several exciting avenues for future research. First, it enables us to build a Conditional Situation Localizer (CSL); a model that outputs a grounded situation conditioned on an input image and a specified region of interest within the image. CSL allows us to query \emph{what} is happening in an image in regards to a specified \emph{query object or region}. This is particularly revealing when entities are involved in multiple situations within an image or when an image consists of a large number of visible entities. Second, we show that such pointed conditioning models enable us to tackle higher order semantic relations amongst activities in images via visual chaining. Third, we show that grounded situation recognition models can serve as effective image retrieval mechanisms that can condition on linguistic as well as visual inputs and are able to retrieve images with the desired semantics.

In summary our contributions include: (i) proposing Grounded Situation Recognition, a task to identify the observed salient situation and ground the corresponding roles within the image, (ii) presenting the \dataset dataset towards building and benchmarking models for this task, (iii) showing that joint structured semantic prediction and grounding models improve both semantic role prediction and grounding by large margins, but also noting that there is still considerable ground for future improvements; (iv) revealing several exciting avenues for future research that exploit grounded situation recognition data to build models for semantic querying, visual chaining, and image retrieval. Our new dataset, code, and trained model weights will be publicly released.

\section{Related Work}

Grounded Situation Recognition is related to several areas of research at the intersection of vision and language and we now present a review of these below. 

\vspace{-5mm}
\subsubsection{Describing Activities in Images. }
While recognizing actions in
videos has been a major focus area~\cite{ucf101,KineticsDataset2017,ActivityNet2015,Sigurdsson2016HollywoodIH,Sigurdsson2018CharadesEgoAL}, describing activities from images has also received a lot of attention (see Gella \etal\cite{analysisActionRecogDatasets17} for a more detailed overview).

Early works~\cite{Ikizler2008RecognizingAF,Gupta2009ObservingHI,Delaitre2010RecognizingHA,Stanford40Actions,Yao2010GroupletAS,Le2013ExploitingLM,everingham2015pascal} framed this as a classification problem amongst a few verbs (running/walking/etc.) or few verb-object tuples (riding bike/riding horse/etc.). More recent work has focused on human object interactions \cite{chao2015hico,tuhoi,hcvrd,ronchi2015describing} with more classes; but the classes are either arbitrarily chosen or obtained by starting with a set of images and then labeling them with actions. Also, the relationships include Subject-Verb-Object triples or subsets thereof. In contrast, the \imsitu{} dataset for situation recognition uses linguistic resources to define a large and more comprehensive space of possible situations, ensuring a fairly balanced datasets despite the large number of verbs (roughly 500) and modeling a detailed set of semantic roles per verb obtained from FrameNet~\cite{framenet}.

Image captioning is another popular setup to describe the salient actions taking place in an image with several datasets \cite{chen2015microsoft,sharma2018conceptual,Agrawal2019nocapsNO} and many recent neural models that perform well \cite{Vinyals2014ShowAT,anderson2018bottom,Karpathy2015DeepVA}. One serious drawback to image captioning is the well known challenge of evaluation which has led to a number of proposed metrics~\cite{meteor,cider,spice,rouge,bleu}; but these problems continue to persist. Situation recognition does not face this issue and has clearly established metrics for evaluation owing to its structured frame output.

Other relevant works include visual sense disambiguation~\cite{Gella2016UnsupervisedVS}, visual semantic role labelling~\cite{gupta2015visual}, and scene graph generation~\cite{visualgenome} with the latter two described in more detail below.

\vspace{-5mm}
\subsubsection{Visual Grounding. }
In contrast to associating full images with actions or captions, past works have also associated regions to parts of captions. This includes \emph{visual grounding} i.e. associating words in a caption to regions in an image and \emph{referring expression generation} i.e. producing a caption to unambiguously describe a region of interest; and there are several interesting datasets here. 

Flickr30k-Entities~\cite{flickr30kentities} is a large dataset for grounded captioning. v-COCO~\cite{gupta2015visual} is more focused on semantic role labeling for human interactions with human groundings, action labels and relevant object groundings. Compared to \dataset{}, the verbs (26 vs 504) and semantic roles per verb (up to 2 vs up to 6) are fewer. HICO-Det~\cite{hicodet} has 117 actions, but they only involve 80 objects, compared to nearly 10,000 objects in \dataset{}. In addition to these human centric datasets, \dataset{} also contains actions by animals and objects.

Large referring expression datasets include RefClef~\cite{ReferItGame}, RefCOCO~\cite{refcoco} and RefCOCO+ collected using a two person game, RefCOCOg collected by standard crowdsourcing and GuessWhat?!~\cite{GuessWhat} that combines dialog and visual grounding.

An all encompassing vision and language dataset is Visual Genome (VG)~\cite{visualgenome} containing scene graphs: dense structured representations for images with objects, attributes, relations, groundings and QA. VG differs from \dataset{} in a few ways. Scene graphs are dense while situations capture salient activities. Also, relations in scene graphs are binary and tend to favor part and positional relations (the top 10 relations in VG are of this nature and cover 66\% of the total) while \dataset{} contains more roles per verb, has 504 verbs drawn from language and has a good coverage of data per verb. Finally, dense annotations are notoriously hard to obtain; and it is well known that VG suffers from missing relations, rendering evaluation tricky.

\vspace{-5mm}
\subsubsection{Situation Recognition Models. }
Yatskar \etal~\cite{imsitu} present a conditional random field model fed by CNN features and extend it with semantic sparsity augmentation \cite{imsitucrf}. Mallya \etal~\cite{imsiturnn} improve the accuracy by using a specialized verb predictor and an RNN for noun prediction. Li \etal~\cite{li2017situation} use Graph Neural Nets to capture joint dependencies between roles. Most recently, Suhail \etal~\cite{suhail2019mixture} achieved state of the art accuracy using attention graph neural nets. Our proposed grounded models build upon the RNN based approach of \cite{imsiturnn} owing to its simplicity and high accuracy; but our methods to combine situation recognition models with detectors can be applied to any of the aforementioned approaches.

\vspace{-5mm}
\subsubsection{Large-Class-Cardinality Object Detection. } 
While most popular object detectors are built and evaluated on datasets~\cite{Lin2014MicrosoftCC,Everingham2009ThePV} with few classes, some past works have addressed the problem of building detectors for thousands of classes. This includes YOLO-9000~\cite{yolo}, DLM-FA \cite{Yang2019Detecting1C}, R-FCN-3000 \cite{Singh2017RFCN3000A3}, and CS-R-FCN \cite{GuoLiAndWang2020}. Our modifications to RetinaNet borrow some ideas from these works.

\section{\tasksmall and \dataset{}}

\vspace{-10mm}
\begin{figure}
\centering
\includegraphics[height=3.4cm]{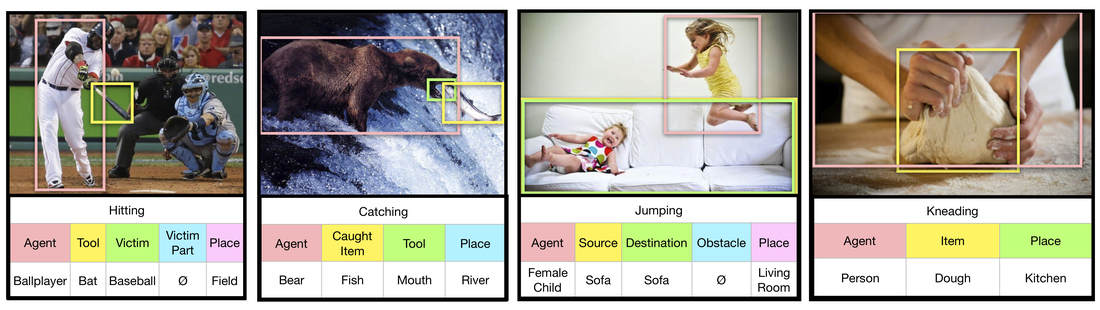}
\vspace{-1em}
\caption{\textbf{Grounded situations from the \dataset{} dataset.} This figure showcases the variability of images, situations and groundings across the dataset. Some challenges seen in this figure are absent roles (first image), animals as agents (second image) contrasting datasets that only focus on human interactions, ambiguity resolution (two female children in the third image), matching groundings for two roles (sofa in the third image) and partial occlusion (person only partially visible in the fourth image)}
\label{fig:teaser2}
\end{figure}
\vspace{-3em}

\subsubsection{Task. }
Grounded Situation Recognition (\tasksmall) builds upon situation recognition and requires one to identify the salient activity, the entities involved, the semantic roles they play and the locations of each entity in the image. The frame representation is drawn from the linguistic resource FrameNet and the visual groundings are akin to bounding boxes produced by object detectors. More formally, given an input image, the goal is to produce three outputs. (a) \textbf{Verb}: classifying the salient activity into one of 504 visually groundable verbs (one in which it is possible to view the action, for example, ‘talking’ is visible, but ‘thinking’ is not). (b) \textbf{Frame}: consists of 1 to 6 semantic role values i.e. nouns associated with the verb (each verb has its own pre-defined set of roles). For e.g., Fig.~\ref{fig:teaser2} shows that ‘kneading’ consists of 3 roles: ‘Agent’, ‘Item’, and ‘Place’. Every image labeled with the verb ‘kneading’ will have the same roles but may have different nouns filled in at each role based on the contents of the image. A role value can also be $\varnothing$ indicating that a role does not exist in an image (Fig.~\ref{fig:teaser2}c). (c) \textbf{Groundings}: each grounding is described with coordinates $[x_1, y_1, x_2, y_2]$ if the noun in grounded in the image. It is possible for a noun to be labeled in the frame but not grounded, for example in cases of occlusion.

\subsubsection{Data. }

\dataset{} builds on top of \imsitu{}~\cite{imsitu}. \dataset{} retains the original images, frame annotations and splits from \imsitu{} with a total of 126,102 images spanning 504 verbs. For each image, there are three frames by three different annotators, with a length between 1 and 6 roles and an average frame length of 3.55.

Bounding-box annotations were obtained using Amazon's Mechanical Turk framework with each role annotated by three workers and the resulting boxes combined by averaging their extents. In total, \dataset{} contains 451,916 noun slots across all images. Of these 435,566 are non-$\varnothing$. Of these 278,336 (63.9\%) have bounding boxes. The missing bounding boxes correspond to objects that are not visible or to `Place' which is never annotated with a bounding box as the location of an action is always the entire image.

\begin{figure*}[h!]
\begin{center}
\includegraphics[height=5cm]{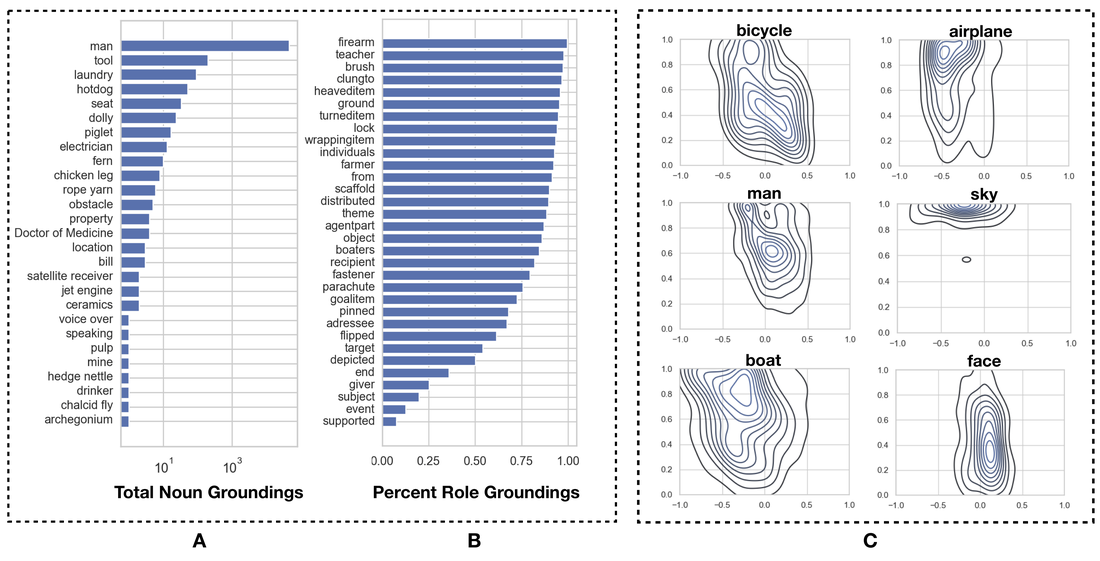}
\end{center}
\vspace{-7mm}
\caption{\textbf{Dataset visualizations.} (A) Number of groundings per noun. Note the log scale and the fact that this only shows a small sample. (B) Frequency with which different roles are grounded in the image. (C) Distribution of grounding scale (y-axis) and aspect ratio (x-axis) conditioned on some nouns}
\label{fig:freqgroundings}
\vspace{-5mm}
\end{figure*}

\dataset{} exhibits a huge variability in the number of groundings per noun (see Fig.~\ref{fig:freqgroundings}a). For instance `man' appears over 100k times while others occur only once. Unlike other detection datasets such as MS-COCO~\cite{chen2015microsoft}, \dataset{} contains a long-tail distribution of grounded objects similar to the real world.  

\vspace{-6mm}

Fig. \ref{fig:freqgroundings}b shows the frequency with which different roles are grounded in the image. Note that, like nouns, roles also have an uneven distribution. Almost all situations are centered around an `Agent' but very few situations use a `Firearm'. This plot shows how often each role is grounded invariant to its absolute frequency. Some roles are much more frequently salient, demonstrating the linguistic frame's ability to capture both concrete and abstract concepts related to situations. Objects filling roles like `Firearm'/`Teacher' are visible nearly every time they are relevant to a situation. However, the noun taking on the role of the `Event' cannot usually be described by a particular object in the image. Only one role (`Place') is never grounded in the image.

Fig.~\ref{fig:freqgroundings}c shows the distribution of grounding scale and aspect ratio for a sample of nouns. Many nouns exhibit high variability across the dataset (1st column), but some nouns have strong priors that may be used by models (2nd column).

\begin{figure*}[h!]
\begin{center}
\includegraphics[height=6cm]{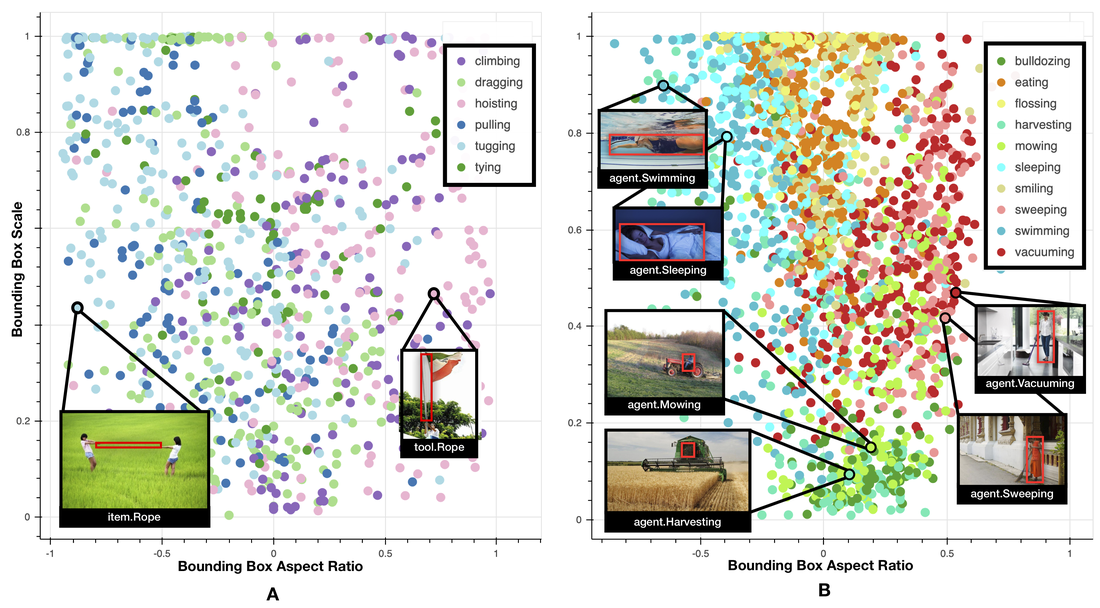}
\end{center}
\vspace{-5mm}
\caption{\textbf{Scale and aspect ratio distributions across nouns and roles.} (A) Every occurrence of the noun `Rope' for verbs - showing that verb gives a strong prior towards the physical attributes of an object. (B) The role `Agent', invariant to the noun - shows priors but also the challenges of the task}
\label{fig:scatterplot}
\vspace{-5mm}
\end{figure*}

Fig. \ref{fig:scatterplot} shows the variability of appearance of groundings across verbs. Fig.~\ref{fig:scatterplot}a indicates the scale and aspect ratio of every occurrence of the noun `Rope' for verbs where this noun occurs at least 60 times. Each point is an instance and the color represents the verb label for the image where that instance appears. A large scale indicates that at least one side of the bounding box is large in relation to the image. A large aspect ratio indicates that the height of the bounding box is much greater than the width. This plot shows that the verb associated with an image gives a strong prior towards the physical attributes of an object. In this case, knowing that a rope appears in a situation with the verb `drag' or `pull', indicates that it is likely to have a horizontal alignment. If the situation is `hoisting' or `climbing' then the rope is likely to have a vertical alignment. 

Fig.~\ref{fig:scatterplot}b shows the scale and aspect ratio of the role `Agent', invariant to the noun, for a variety of verbs. The clustering of colors in the plot indicates that the verb gives a strong prior to the size and aspect ratio of the `Agent'. However, this also demonstrates the non-triviality of the task. This is especially evident in the images depicting the agent for `Mowing' compared to `Harvesting'. While knowing the verb gives a strong indication as to the appearance of the `Agent', it is not trivial to distinguish between the two verbs given just the agent. The correlation between object appearance and actions demonstrates the importance of combining situation understanding with groundings, but we must still maintain the entire context to complete the task.  

\vspace{-5mm}

\vspace{-5mm}
\section{Methods}
\label{sec:methods}
\vspace{-2mm}

Grounded situation recognition involves recognizing the salient situation and grounding the associated role values via bounding boxes; indicating that a model for this task must perform the roles of situation recognition and object detection. We present a novel method Joint Situation Localization (JSL) with a strong baseline, the Independent Situation Localization (ISL). 

\vspace{-5mm}
\subsubsection{Situation Recognition Model. }
The proposed ISL and JSL models represent techniques to combine situation recognition and detection models and can be applied to all past situation recognition models. In this work, we select the RNN without fusion from \cite{imsiturnn} since: (i) it achieves a high accuracy while having a simple architecture (Table~\ref{table:devresults}). (ii) We were able to \emph{upgrade} it with a reimplementation, new backbone, label smoothing, and hyper-parameter tuning resulting in huge gains (Table~\ref{table:devresults}) over the reported numbers, beating graph nets~\cite{li2017situation} and much closer to attention graph nets~\cite{suhail2019mixture} (the current state-of-the-art on \imsitu). (iii) Code and models for attention graph nets are not released, rendering reproducibility challenging, especially given the complexity of the method.

\begin{figure*}[t!]
\begin{center}
\includegraphics[width=\textwidth]{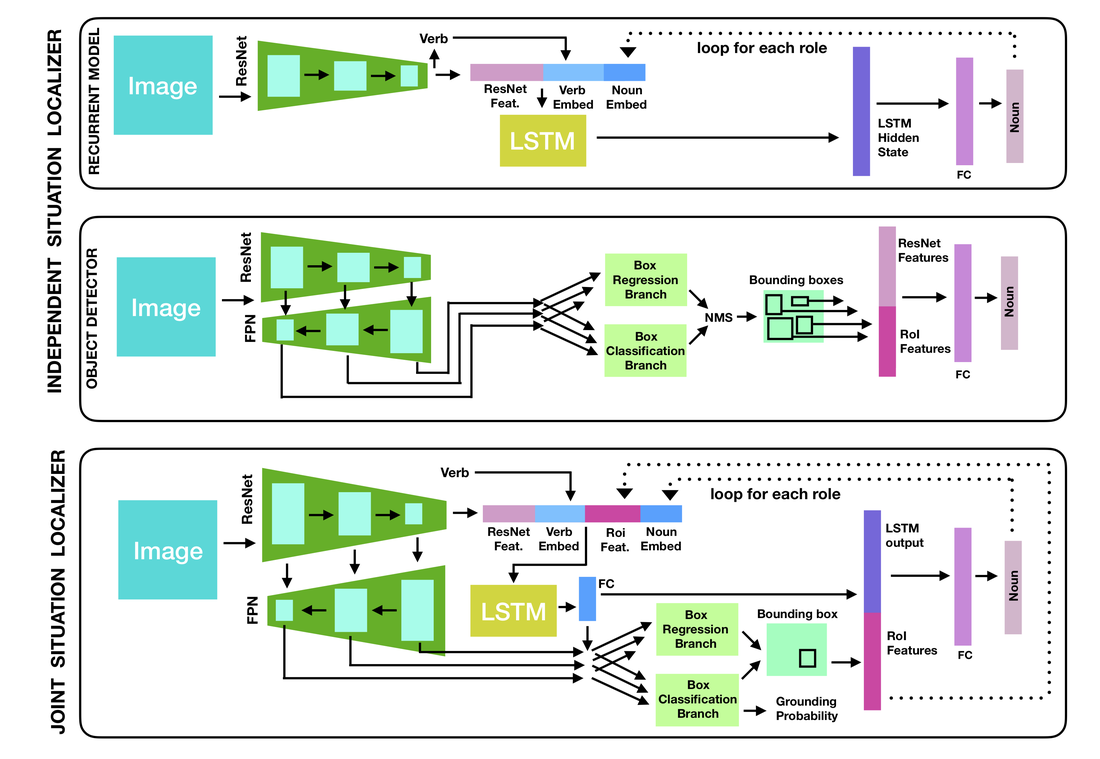}
\end{center}
\vspace{-5mm}
\caption{Model schematics for the proposed ISL and JSL models}
\vspace{-5mm}
\label{fig:model}
\end{figure*}

As in the top of Fig.~\ref{fig:model}, ResNet-50 embeddings are used for verb prediction and then an LSTM~\cite{lstm} sequentially predicts the noun for each role in the frame. The order of the roles is dictated by the dataset. The loss is a sum of cross entropy with label smoothing on the predicted nouns, and cross entropy on the verb.

\vspace{-5mm}
\subsubsection{Large-Class-Cardinality Object Detection. }
We use a modified version of RetinaNet~\cite{retinanet} as our baseline detector within the proposed models. RetinaNet is a single stage detector with a Feature Pyramidal Network (FPN)~\cite{fpn} for multi-scale features, multiple anchors to account for varied aspect ratios and two heads: a classification head that assigns each anchor to a class and a regression head that modifies the anchor to better localize any object in that location. RetinaNet does not scale well to the 10k classes in \dataset{} out of the box.

We make 3 modifications to RetinaNet to scale it to 10,000 classes, as seen in the middle of Fig.~\ref{fig:model}. (i) \textbf{Objectness}: instead of each anchor predicting a score for each class, each anchor now predicts an ``objectness score". Non-Maximum Suppression (NMS) is performed on the boxes, the top 100 are chosen and featurized using RoI Align~\cite{fasterrcnn}. These local features combined with global ResNet features are classified into the $\sim$10,000 noun categories. The resulting memory savings are huge. In RetinaNet, the classification branch output tensor has dimensions $\sum_{i=1}^n(W_i \times H_i \times A \times K)$ where $W_i,H_i$ indicate the spatial dimensional of the features for the $i^{th}$ output of the FPN, $A$ indicates the number of anchor boxes, and $K$ indicates the number of classes. This does not fit on a single TITAN RTX GPU for $K = 10,000$ for any reasonable batch size. In contrast, our modification reduces the tensor dimension to $\sum_{i=1}^n(W_i \times H_i \times A \times P)$ where $P$ is the number of image regions we consider and is set to 100. With these modifications we are able to train with a batch size of 64 on 4 TITAN RTX GPUs with 24GB of memory. (ii) \textbf{Drop fine scale}: we exclude the finest grain of features from FPN since anchors at this scale do not overlap with a significant portion of our data leading to computation savings. (iii) \textbf{Anchor selection}: anchor box aspect ratios are assigned using aspect ratio clustering on our training data, as in \cite{yolo}. As in \cite{retinanet}, we use a focal loss for classification and an $L_1$ loss for regression, with a binary cross entropy loss for noun prediction.

\vspace{-5mm}
\subsubsection{Independent Situation Localizer (ISL). } 
The ISL independently runs the situation recognizer and detector and combines their results. The RNN model produces a prediction for each noun in the frame. The detector obtains a distribution over all possible object categories for each of the top 100 bounding boxes. Then for each noun in the frame, we assign the grounding with the highest score for that noun. This allows an object that is assigned to one class by the detector to eventually get assigned to another class as long as the score for the latter class is high enough. If all of the box scores for a noun are below a threshold or the role is `Place', it is considered ungrounded. 

\vspace{-5mm}
\subsubsection{Joint Situation Localizer (JSL). }
We propose JSL as a method to simultaneously classify a situation and locate objects in that situation. This allows for a role's noun and grounding to be conditioned on the nouns and groundings of previous roles and the verb. It also allows features to be shared potential patterns between nouns and positions (like in Fig.~\ref{fig:scatterplot}) to be exploited. We refer the reader to the appendix and our code for model details, but point out key differences between JSL and ISL here. 

JSL (shown in the bottom of Fig.~\ref{fig:model}) uses similar backbones as ISL but with key differences: (i) rather than predicting localization for every object in the image at the same time (as is done in object detection), JSL predicts the location of the objects recurrently (as is done for predicting nouns in situation recognition). (ii) In contrast to the RNN model in ISL, the LSTM accepts as input the verb embedding, global ResNet features of the image, embedding of the noun predicted at the previous time step and local ResNet features of the bounding box predicted at the previous time step. (iii) In contrast to the detector in ISL, the localization is now conditioned on the situation and current role being predicted. In ISL, FPN features feed directly into the classification and regression branches, but in JSL the FPN features are combined with the LSTM hidden state and then fed in. (iv) The JSL also uses the classification branch to produce an explicit score indicating the likelihood that an object is grounded. (v) Only one noun needs to be localized at each time step, which means that only anchor boxes relevant to that one grounding will be marked as positive during training, given that the noun is visible and grounded in the training data.

The loss includes focal loss and $L_1$ loss from the detector and cross entropy loss for the grounding and verb. Additionally, we use cross entropy with label smoothing for the noun loss, and sum this over all three annotator predictions. This results in the following total loss:
$\mathcal{L} = L_1(reg) + FL_{0.25,2}(class) + CE(verb) + CE(ground) + \sum_{i=1}^{3} CE_{0.2}({noun_i})$

Similar to previous works \cite{imsiturnn,suhail2019mixture}, we found that using a separate ResNet backbone to predict the verb achieved a boost in accuracy. However, the JSL architecture with this additional ResNet backbone still maintains the same number of parameters and ResNet backbones as the ISL model.

\definecolor{Gray}{gray}{0.83}
\definecolor{Gray_2}{gray}{0.9}
\definecolor{yellow_1}{RGB}{255, 255, 185}
\definecolor{green_1}{RGB}{210, 255, 210}

\vspace{-3mm}
\section{Experiments}
\vspace{-2mm}

\subsubsection{Implementation Details. }

ISL and JSL use two ResNet-50 backbones and maintain an equal number of parameters ($\sim$108 million). We train our models via gradient descent using the Adam Optimizer \cite{KingmaAndBa2015} with momentum parameters of $\beta=(0.9, 0.999)$. We use 4 24GB TITAN RTX GPUs for approximately 20 hours. For comprehensive training and model details, including learning rate schedules, batch sizes, and layer sizes, please see our appendix.

\vspace{-5mm}
\subsubsection{Metrics. }
We report five metrics, three standard ones from prior situation recognition work: (i) \textbf{verb} to measure verb prediction accuracy, (ii) \textbf{value} to measure accuracy when predicting a noun for a given role, (iii) \textbf{value-all} to measure the accuracy in correctly predicting all nouns in a frame simultaneously; and introduce two new grounding metrics: (iv) \textbf{grounded-value} to measure  accuracy in predicting the correct noun and grounding for a given role. A grounding is considered correct if it has an IoU of at least 0.5 with the ground truth. (v) \textbf{grounded-value-all} to measures how frequently both the noun and the groundings are predicted correctly for the entire frame. Note that if a noun does not have a grounding, the model must also predict this correctly. All these metrics are calculated for each verb and then averaged across verbs so as to not unfairly bias this metric toward verbs with more annotations or longer semantic frames. 

Since these metrics are highly dependent on verb accuracy, they have the potential to obfuscate model differences with regards to noun prediction and grounding. Hence we report them in 3 settings: \textbf{Ground-Truth-Verb}: the ground truth verb is assumed to be known. \textbf{Top-1-Verb}: \emph{verb} reports the accuracy of the top 1 predicted verb and all noun and groundings are considered incorrect if the verb is incorrect. \textbf{Top-5-Verb}: \emph{verb} corresponds to the top-5 accuracy of verb prediction. Noun and grounding predicitons are taken from the model conditioning on the correct verb having been predicted.

\begin{table*}
\centering
\caption{Evaluation of models on the \dataset{} dev set. * indicates our implementation. \colorbox{yellow!30}{Yellow rows} indicate the base RNN model architecture with numbers from the paper. \colorbox{green!30}{Green} shows the upgraded version of this RNN model used in our proposed models}
\vspace{-2mm}

\resizebox{\textwidth}{!}{
\begin{tabular}{|l*{14}{|c}|}
\hline
\multirow{3}{2em}{~\\~\\Method} & \multicolumn{5}{|c|}{top-1 predicted verb} & \multicolumn{5}{|c|}{top-5 predicted verbs} & \multicolumn{4}{|c|}{ground truth verbs} \\
\cline{2-15}
&  & & & grnd & grnd&  &  &  & grnd  & grnd&  & & grnd  & grnd   \\
& verb & value & value-all &  value &  value-all & verb & value & value-all &  value &  value-all & value & value-all &  value &  value-all  \\\hline
\multicolumn{15}{|c|}{\textbf{Prior Models for Situation Recognition}}\\\hline
CRF \cite{imsitu}  & 32.25 & 24.56 & 14.28 & - & - & 58.64 & 42.68 & 22.75 & - & - & 65.90 & 29.50 & - & -\\\cline{2-15}
CRF+Aug \cite{imsitucrf} & 34.20 & 25.39 & 15.61 & - & - & 62.21 & 46.72 & 25.66 & - & - & 70.80 & 34.82 & - & -\\\cline{2-15}
\rowcolor{yellow_1}
RNN w/o Fusion\cite{imsiturnn} & 35.35 & 26.80 & 15.77 & - & - & 61.42 & 44.84 & 24.31 & - & - & 68.44 & 32.98 & - & -  \\\cline{2-15}
RNN w/ Fusion\cite{imsiturnn} & 36.11 & 27.74 & 16.60 & - & - & 63.11 & 47.09 & 26.48 & - & - & 70.48 & 35.56 & - & - \\\cline{2-15}
GraphNet \cite{li2017situation} & 36.93 & 27.52 & 19.15 & - & - & 61.80 & 45.23 & 29.98 & - & - & 68.89 & 41.07 & - & - \\\cline{2-15}
Kernel GraphNet\cite{suhail2019mixture} & \textbf{43.21} & \textbf{35.18} & \textbf{19.46} & - & - & \textbf{68.55} & \textbf{56.32} & \textbf{30.56} & - & - & \textbf{73.14} & \textbf{41.48} & - & -  \\\hline
\multicolumn{15}{|c|}{\textbf{RNN based models}}\\\hline
\rowcolor{yellow_1}
RNN w/o Fusion \cite{imsiturnn} & 35.35 & 26.80 & 15.77 & - & - & 61.42 & 44.84 & 24.31 & - & - & 68.44 & 32.98 & - & - \\\cline{2-15}
\rowcolor{green_1}
Updated RNN* & 38.83&30.47&18.23& - & - &65.74&50.29&28.59& - & - &72.77&37.49&-&- \\\cline{2-15}
ISL* & 38.83&30.47&18.23&22.47&7.64&65.74&50.29&28.59&36.90&11.66&72.77&37.49&52.92&15.00 \\\cline{2-15}
JSL* & \textbf{39.60} & \textbf{31.18} & \textbf{18.85} & \textbf{25.03} & \textbf{10.16} & \textbf{67.71} & \textbf{52.06} & \textbf{29.73} & \textbf{41.25} & \textbf{15.07} & \textbf{73.53} & \textbf{38.32} & \textbf{57.50} & \textbf{19.29} \\\hline
\end{tabular}}
\label{table:devresults}
\vspace{-3mm}
\end{table*}

\begin{table}[h!]
\setlength{\tabcolsep}{3pt}
\centering
\caption{Evaluation of models on the \dataset{} test set. * indicates our implementation}
\vspace{-2mm}
\resizebox{\textwidth}{!}{%
\begin{tabular}{|l*{14}{|c}|}
\hline
\multirow{3}{2em}{~\\~\\ Method} & \multicolumn{5}{|c|}{top-1 predicted verb} & \multicolumn{5}{|c|}{top-5 predicted verbs} & \multicolumn{4}{|c|}{ground truth verbs} \\\cline{2-15}
&  & & & grnd & grnd&  &  &  & grnd  & grnd&  & & grnd  & grnd  \\
& verb & value & value-all &  value &  value-all & verb & value & value-all &  value &  value-all & value & value-all &  value &  value-all\\\hline
\multicolumn{15}{|c|}{\textbf{RNN based models}}\\\hline
Updated RNN* \hphantom{-------------} & 39.36 & 30.09&18.62& - & - &65.51&50.16&28.47& - & -  &  72.42 & 37.10 & - & -\\\cline{2-15}
ISL* & 39.36 & 30.09&18.62&22.73&7.72&65.51&50.16&28.47&36.60&11.56 &  72.42 & 37.10 & 52.19 & 14.58\\\cline{2-15}
JSL* & \textbf{39.94} & \textbf{31.44} & \textbf{18.87} & \textbf{24.86} & \textbf{9.66} & \textbf{67.60} & \textbf{51.88} & \textbf{29.39} & \textbf{40.60} & \textbf{14.72} & \textbf{73.21} & \textbf{37.82} & \textbf{56.57} & \textbf{18.45}\\\hline
\end{tabular}}
\label{table:testresults}
\vspace{-6mm}
\end{table}

\vspace{-5mm}
\subsubsection{Results. }
The top section of Table \ref{table:devresults} shows past \imsitu models for the dev set while the lower section illustrates the efficacy of jointly training a model for grounding and situation recognition. The yellow rows indicate the base RNN model used in this work and the green row shows the large upgrades to this model across all metrics. ISL achieves reasonable results, especially for ground truth verbs. However, JSL improves over ISL across every metric while using an equal number of parameters. This includes substantial improvements on all grounding metrics (ranging from relative improvements of 8.6\% for \emph{Ground-Truth-Verb--ground-value} to 32.9\% for \emph{Top-1-Verb--grounded-value-all}). 

The ability to improve across both the grounding and non-grounding scores demonstrate the value in combining grounding with the task of situation recognition. Not only can the context of the situation improve the models ability to locate objects, but locating these objects improves the models ability to understand them. This is further emphasized by the models ability to predict the correct noun under the \emph{GroundTruthVerb} setting. 

\begin{figure*}[!t]
\begin{center}
\includegraphics[width=\textwidth]{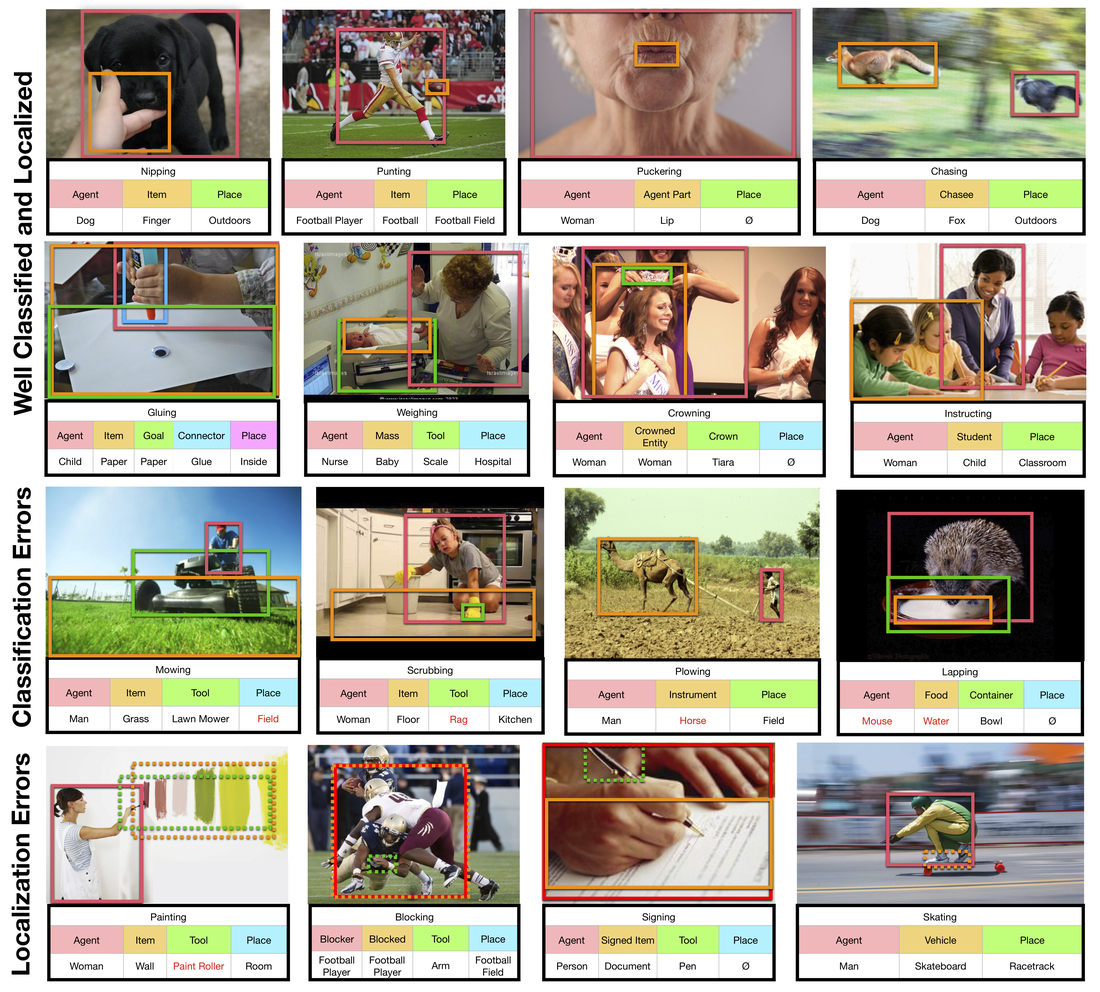}
\end{center}
\vspace{-5mm}
\caption{\textbf{Qualitative results for the proposed JSL model.} First two rows show examples with correctly classified situations and detected groundings; and demonstrates the diversity of situations in the data. Third row shows classification errors. Note that some of them are perfectly plausible answers. Fourth row shows incorrect groundings; some of which are only partially wrong but get counted as errors nonetheless}
\label{fig:qualitative}
\vspace{-5mm}
\end{figure*}

Importantly, in spite of using the simpler RNN based backbone for situation recognition, JSL achieves state of the art numbers on the GroundTruthVerb-Value metric, beating the more complex Kernel GraphNet model demonstrating the benefits of joint prediction. This indicates that further improvements may be obtained by incorporating better backbones. Additionally, it is interesting to note that the model achieves this high \emph{value} even though it does not achieve state of the art in \emph{value-all}. This indicates another potential benefit of the model. While more total frames contain a mistake, JSL is still able to recover some partial information and recognize more total objects. One explanation is that grounding may contribute to the ability to have a \emph{partial} understanding of more complicated images where other models fail completely. Finally, test set metrics are shown in Table \ref{table:testresults} and qualitative results in Fig~\ref{fig:qualitative}.

\vspace{-3mm}
\section{Discussion}
\label{sec:discussion}

\vspace{-2mm}
Grounded situation recognition and \dataset{} open up several exciting directions for future research. We present initial findings for some of these explorations.

\vspace{-4mm}
\begin{figure*}[h!]
\begin{center}
\includegraphics[width=\textwidth]{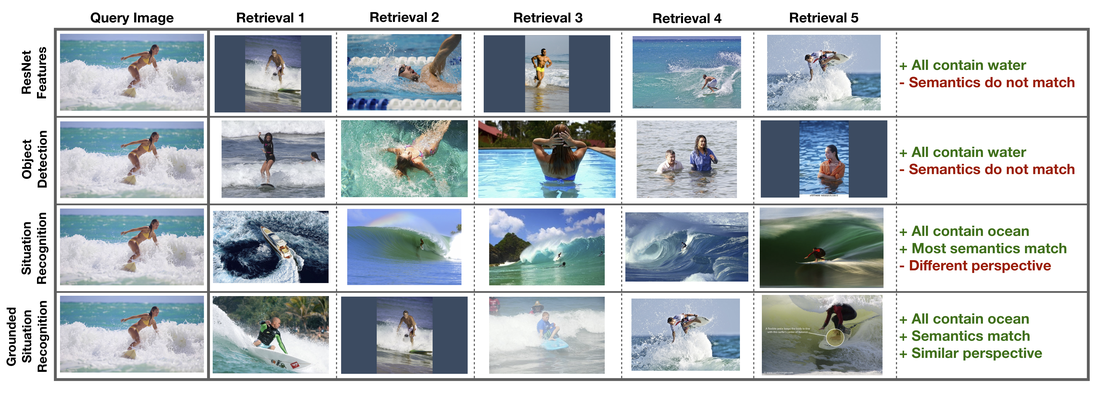}
\end{center}
\vspace{-5mm}
   \caption{\textbf{Qualitative results for semantic image retrieval}. For the query figure of a surfer in action, ResNet and Object Detection based methods struggle to match the fine semantics. Grounded situation based retrieval leads to the correct semantics with matching viewpoints}
\label{fig:nn}
\vspace{-5mm}
\end{figure*}

\vspace{-5mm}
\subsubsection{Grounded Semantic aware Image Retrieval. }
Over the past few years, large improvements have been obtained in content based image retrieval (CBIR) by employing visual representations from CNNs~\cite{Babenko2014NeuralCF,Radenovi2016CNNIR,Razavian2014VisualIR,Gordo2016DeepIR,Yang2018EfficientIR}. CNN features work well particularly when evaluated on datasets requiring instance retrieval~\cite{Philbin2007ObjectRW} or category retrieval~\cite{imagenet}, but unsurprisingly do not do well when the intent of the query is finding a matching situation. We perform a small study for image retrieval using the dev set in \dataset{}. We partition this set into a query set and a retrieval set and perform retrieval using four representations: (i) ResNet-50 embeddings, (ii) bag of objects obtained from our modified RetinaNet object detector, (iii) situations obtained from our baseline RNN model, and (iv) grounded situations obtained from JSL. Details regarding the setup and distance functions for each are presented in the appendix.

Fig.~\ref{fig:nn} shows a qualitative result. Resnet-50 retrieves images that look similar (all have water) but have the wrong situations. The same goes for object detection. Situation based retrieval gets the semantics correct (most of the retrieved images contain surfing). Grounded situations provide the additional detail of not just similar semantics but also similar arrangement of objects, since the locations of the entities are also taken into account. Furthermore, the proposed method also produces explainable outputs via the grounded situations in the retrieved images; arguably more useful than CBIR explanations via heatmaps~\cite{Dong_2019_CVPR_Workshops}. This approach can also be extended to structured queries (obtained from text) and a mix of text and image based queries.

\begin{figure*}[h!]
\begin{center}
\includegraphics[width=\textwidth]{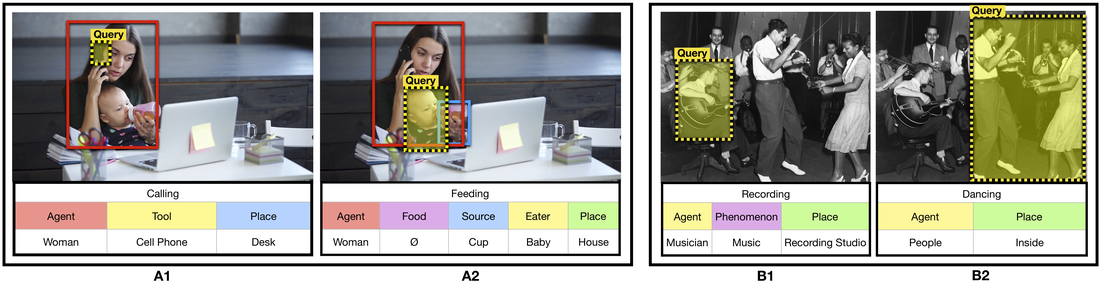}
\end{center}
\vspace{-5mm}
   \caption{\textbf{Qualitative results using the Conditional Situation Localizer.} A1 \& A2: The woman is taking part in multiple situations with different entities in the scene. These situations are invoked via different queries. B1 \& B2: Querying the person with a guitar vs querying the group of people also reveals their corresponding situations}
\label{fig:conditional}
\vspace{-5mm}
\end{figure*}

\vspace{-5mm}
\subsubsection{Conditional Grounded Situation Recognition. }
JSL accepts the entire image as an input and produces groundings for the salient situation. But images may contain entities in multiple situations (a person \emph{sitting} and \emph{discussing} and \emph{drinking coffee}) or multiple entities. Conditioning on a localized object or region can enable us to \textit{query} an image regarding the entity of interest. Note that a query entity may be an actor (\emph{what is this person doing?}) or an object (\emph{What situation is this object involved in?}) or a location (\emph{What is happening at this specific location?}) in the scene. A small modification to JSL results in a Conditional Situation Localizer (CSL) model (details in appendix), which enables this exploration. Fig.~\ref{fig:conditional}a shows that a query box around the cellphone invokes \emph{calling} while the baby invokes \emph{feeding}. Fig.~\ref{fig:conditional}b shows that a query box may have 1 or more entities within it.

\begin{figure*}[h!]
\begin{center}
\includegraphics[width=\textwidth]{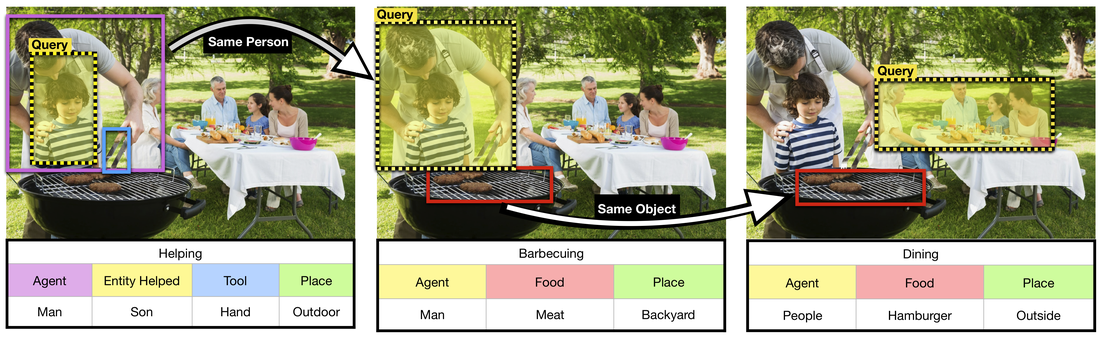}
\end{center}
\vspace{-5mm}
   \caption{\textbf{Grounded semantic chaining.} When a person looks at this image, they may infer several things. A father is teaching his son to use the grill. They are barbecuing some meat with the intent of feeding friends and family who are sitting at a nearby table. Using the conditional localizer followed by spatial and semantic chaining produces situations and relationships-between-situations. These are shown via colored boxes, text and arrows. Conditional inputs are shown with dashed yellow boxes. Notice the similarity between the higher level semantics output by this chaining model and the inferences about the image that you may draw}
\label{fig:chaining}
\vspace{-11mm}
\end{figure*}

\subsubsection{Grounded Semantic Chaining. }
Pointed conditional models such as CSL, when invoked on a set of bounding boxes (obtained via object detection), enable us to chain together situations across multiple parts of image. While a situation addresses local semantics, chaining of situations enables us to address higher order semantics across an entire image. Visual chaining can be obtained using spatial and semantic proximity between groundings in different situations. While scene graphs are a formalism towards this, they only enable binary relations between entities; and this data in Visual Genome~\cite{visualgenome} has a large focus on part and spatial relations. Since \dataset{} contains a diverse set of verbs with comprehensive semantic roles, visual chains obtained by CSL tend to be very revealing and are an interesting direction to pursue in future work. Fig.~\ref{fig:chaining} shows an interesting example of querying multiple persons and then chaining the results using groundings, revealing: a man is helping his son while barbecuing meat and the people are dining on the hamburger that is being grilled by the man.

\vspace{-5mm}
\subsubsection{Conclusion}
\vspace{-5mm}

We introduce Grounded Situation Recognition (GSR) and the SWiG dataset. Our experiments reveal that simultaneously predicting the semantic frame and groundings results in huge gains over independent prediction. We also show exciting directions for future research.

\pagebreak
\bibliographystyle{splncs04}
\bibliography{egbib.bib}
\clearpage
\pagebreak
\appendix

\begin{center}

\begin{Large}
\textbf{Appendix}
\end{Large} 

\end{center}

\noindent Here we provide a more detailed explanation of methods introduced in this work and provide additional qualitative results demonstrating the efficacy of our proposed model. In Section~\ref{sec:SIR} we discuss the details of Semantic Image Retrieval as mentioned in Section~\ref{sec:discussion}. In Section~\ref{sec:imp} we provide the implementation details of our baseline model (ISL) and proposed model (JSL). In Section~\ref{sec:csl} we discuss the model changes we make to JSL in order to create the Conditional Situation Localizer as discussed in Section~\ref{sec:discussion}. Finally, in Section~\ref{sec:qual} we provide qualitative results comparing the localization of ISL and JSL as well as qualitative results visualizing the situations generated for the top-5 verbs predicted by JSL.

\section{Semantic Image Retrieval}
\label{sec:SIR}

\begin{figure*}
\centering
\includegraphics[width=\textwidth]{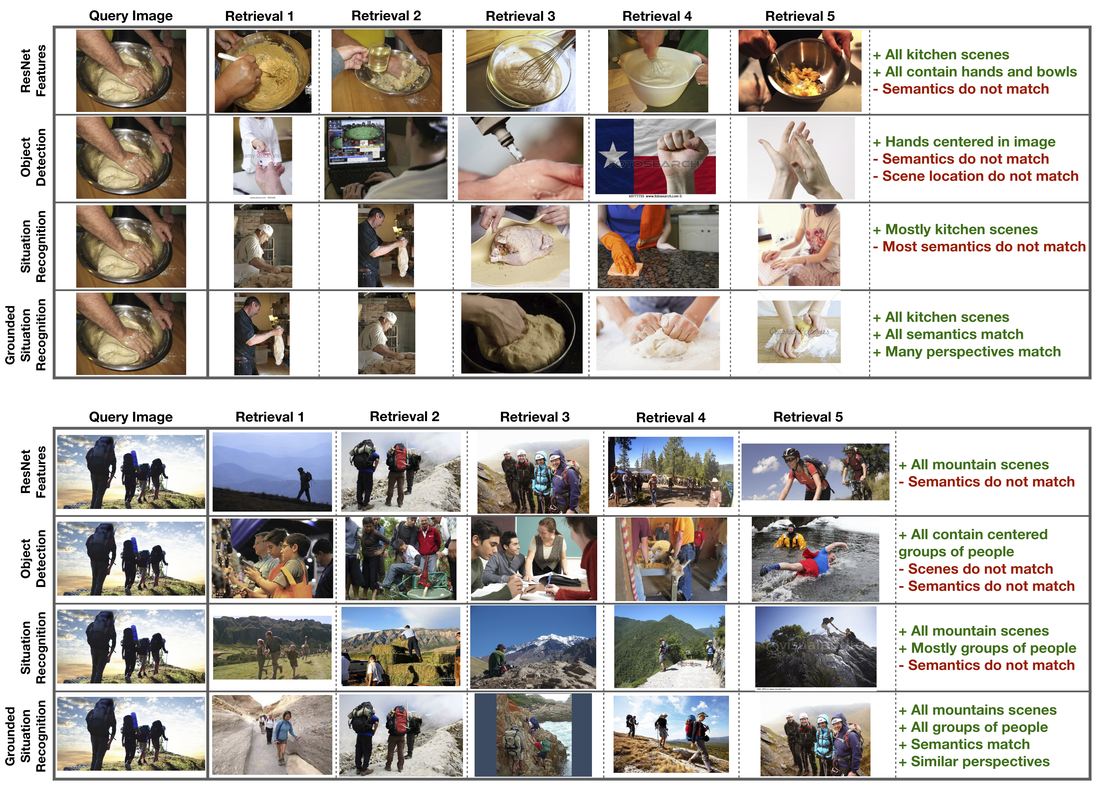}
\caption{\textbf{Additional qualitative results for semantic image retrieval}. For the query figure of a baker kneading dough or multiple hikers walking, ResNet and Object Detection based methods struggle to match the semantics of the image. Grounded situation based retrieval leads to the correct semantics with matching viewpoints}

\label{fig:nn_2}
\end{figure*}

In Fig. \ref{fig:nn} and Fig. \ref{fig:nn_2} we show qualitative examples of semantic image retrieval implemented with nearest neighbor computations and a collection of different similarity functions. In particular, we divide our validation set into a query set (1008 images, 2 images per verb) and search set (24192 images, 48 images per verb). For each of the images in our query set, we compute the similarity of the query image with all images in search set and save the top-5 most similar images to the query. We now describe how we compute image similarity using ResNet-50 features, bag-of-words object detections, situation predictions, and grounded situation predictions.

\subsection{ResNet-50}

We compute a featurization of each image using a ResNet-50 model pretrained on the ImageNet dataset. Similarity between images is then computed as the negative of the L2 distance between these featurizations (so that images with nearer featurizations are more similar).

\subsection{Object Detections}

For each image $I$ we compute object detections using the modified RetinaNet described in Section~\ref{sec:methods}. We find these detections by computing the maximum likelihood category for each box. If the logits corresponding to probability of the maximum category is greater than -1 we consider it a valid detection. To prevent multiple detections of the same object we use NMS to remove any overlapping boxes of the same object category. We save the predicted class labels $\{c^I_1, ..., c^I_{N_I}\}$ and bounding-boxes $\{b^I_1, ..., b^I_{N_I}\}$. Similarity between two images $I,J$ is then computed as
\begin{align}
    \text{ObjSim}(I, J) = \frac{1}{N}\sum_{i=1}^N\max \{1_{[c^I_i=c^J_j]}\cdot (1 + \text{IoU}(b^I_i, b^J_j))\mid 1\leq j\leq M\}
\end{align}
so that $\text{ObjSim}(I, J)$ will be maximal when the objects detected in $I$ have the same classes and bounding-boxes as those in $J$.

\subsection{Situation Recognition}

For each image $I$ in our validation set, we compute $v^I_1,...,v^I_5$ the top-5 predicted activities (verbs) associated with $I$. For each of these verbs $v^I_i$, we additionally predict the entities associated with the roles of that verb, $e^I_{i,1}, ..., e^I_{i, N_{v^I_i}}$. We then compute the situation similarity between two images $I,J$ as
\begin{align}
    \text{SitSim}(I, J) = \max\{\frac{1_{[v^I_i = v^J_j]}}{i\cdot j\cdot N_v}\sum_{k=1}^{N_{v^I_i}} 1_{[e^I_{i,k}=e^J_{j,k}]} \mid 1\leq i,j\leq 5\}.
\end{align}
Notice that $\text{SitSim}(I, J)$ is only non-zero if there is at least one verb shared in the top-5 verb predictions of $I$ and $J$. Moreover, the similarity will be at its maximum value of 1 if any only if both $I$ and $J$ have the same top-1 verb and, for that verb, all predicted entities (conditioned on that top-1 verb) for both images are the same.

\subsection{Grounded Situation Recognition}

As above, we have, for an image $I$ top-5 verb predictions $v^I_1,...,v^I_5$ and entity predictions $\{e^I_{i,k_i}\mid 1\leq i\leq 5,\ 1\leq k_i\leq N_{v^I_i}\}$. For grounded situation predictions we also have, for each entity $e^I_{i,k}$ a bounding-box prediction $b^I_{i, k}$. We then compute similarity between two images $I,J$ as 
\begin{align}
    \text{Gr}&\text{SitSim}(I, J) \nonumber \\
    &= \max\{\frac{1_{[v^I_i = v^J_j]}}{i\cdot j\cdot N_v}\sum_{k=1}^{N_{v^I_i}} 1_{[e^I_{i,k}=e^J_{j,k}]}\cdot (1 + \text{IoU}(b^I_{i,k}, b^J_{j,k})) \mid 1\leq i,j\leq 5\}.
\end{align}
Notice that $\text{GrSitSim}$ is nearly identical to $\text{SitSim}$ except that $\text{GrSitSim}$ will be larger when predicted entities have similar bounding boxes, as measured by their intersection over union.

\section{Implementation Details}
\label{sec:imp}
\subsection{RNN}
\subsubsection{Architecture}
We use a ResNet-50 backbone pretrained on ImageNet. The embedding size for nouns is 512 and the embedding size for verbs is 256. We use a single layer LSTM as the the RNN with a hidden size of 1024 and an input size of 2816 (2048 image features, 512-dimensional embedding of the previous noun, 256-dimensional embedding of the verb). The LSTM is initialized with orthogonal weights. The 512-dimensional noun vector is initialized with zeros for the first noun prediction. The LSTM predicts a sequence length of 6 as this is the maximum length frame. Frames with less than this length are padded to length 6. The ground truth verb embedding is used as input to the LSTM for all of training, as incorrect verb predictions are always marked as having incorrect noun predictions, so there is no benefit to training with incorrect verb predictions. 

\subsubsection{Training} We train the RNN using the Adam Optimizer \cite{KingmaAndBa2015} with $\beta=(0.9, 0.999)$. The initial learning rate is set to 1e-4 which is decreased by a factor of 10 at epoch 12 and 24. Additionally, we begin training by freezing the ResNet weights and only begin to propagate the gradients through ResNet at epoch 14. We train with a batch size of 32 for 100 epochs, which takes ~40 hours on one 12GB TITAN V GPU and use the weights from the best performing epoch. 

\subsection{Object Detector}
\subsubsection{Architecture}
The majority of this architecture is unchanged from the original RetinaNet architecture. We ResNet-50 backbone pretrained on ImageNet. The majority of the differences from the original RetinaNet take place in the adjustments to the network which allow for detection of 10,000 categories. As mentioned in in Section~\ref{sec:methods}, we adjust the network by predicting the likelihood that each anchor box contains an object, rather than predicting a distribution over all object categories for each anchor box. We then perform NMS to remove low scoring boxes which have a high overlap with other boxes. We take the top 100 boxes most likely to contain an object and obtain the features corresponding to these boxes in the final spatial layer of ResNet using RoI align. We then linearly transform the feature vectors into a vector the size of the noun vocabulary to obtain a predicted distribution. For training, if these boxes overlap with a ground truth annotation with an IoU of at least 0.5, they are labeled will all the categories attributed to the ground truth box. A ground truth box may have multiple categories as there are multiple annotators. If it does not overlap with any ground truth box it is not labeled with any category. If it overlaps with multiple ground truth boxes, we duplicate the predicted box and each one is considered to overlap with one ground truth box. We then use binary cross entropy on these labels and the predicted distribution. When combining the RNN output and RetinaNet outputs, a box is assigned to a noun category if it has the highest predicted value for that noun category out of all 100 boxes. If none of the boxes reach a certain threshold for that noun category, then the noun is label as ungrounded in the image. We tune this threshold to be -4 for our model, so if none of the logits are above this value for the desired category, it is ungrounded. 

\subsubsection{Training}
We train with a batch size of 64 using the Adam Optimizer \cite{KingmaAndBa2015} with $\beta=(0.9, 0.999)$. We use a learning rate of 1e-4 for all of training. We train until convergence and then use the weights from the epoch (26) which achieved the highest accuracy on the dev set. We train the network for $\sim$72 hours on eight 12GB TITAN V GPUs. Despite the modifications we made to the RetinaNet model, training is still relatively slow as we must still perform 100 classifications for every image.

\subsection{JSL}
\subsubsection{Architecture}
As with the RNN, we use a single layer LSTM with hidden size 1024, noun embeddings of size 512 and verb embeddings of size 256. We initialize the ResNet-50 backbone with imagenet weights and initialize the LSTM with orthogonal weights. Additionally we pad shorter frames to be of length 6 and label all of the pad symbols to be ungrounded. Like the RNN model, we always use the embedding of the ground truth verb during training, as the nouns are always considered incorrect if the verb prediction is incorrect so there is no benefit to training with the incorrect verb prediction. Additionally, for the first 5 epochs of training, we use the ground truth bounding boxes when obtaining the local features for noun classification and we use the previous ground truth noun when embedding the previous noun for the LSTM. When combining the output of the LSTM with the features from the FPN before the classification and regression branches (see Figure~\ref{fig:model}) we concatenate the FPN features with a linear projection of size 256 of the hidden state of the LSTM. Additionally we concatenate an element wise product of these two vectors, resulting in a final input vector with a channel dimension of 768.  

\subsubsection{Training}
We train with a batch size of 64 using the Adam Optimizer \cite{KingmaAndBa2015} with $\beta=(0.9, 0.999)$. We use an initial learning rate of 6e-4 which we decrease by a factor of 10 at epochs 10 and 20. Like with the RNN, we begin by freezing the ResNet weights and only begin to propagate the weights to the ResNet backbone at epoch 12. We train until convergence and use the weights which have the highest performance on the validation set (epoch 27). Training takes $\sim$20 hours on four 24GB TITAN RTX GPUs. 

\subsubsection{Verb Prediction Network}
As mentioned in Section~\ref{sec:methods}, we find using a separate network to predict the verb increases the accuracy of verb prediction, while keeping the total number of parameters equal to that of the independent model. To train the verb classifier, we use a ResNet backbone with a linear layer on top of the final feature vector of size 2048, just after the final average pooling. We use the Adam Optimizer with an initial learning rate of 1e-4 which we decrease by a factor of 10 at epoch 18. We train just the final linear layer for the first 5 epochs, then just the linear layer and final block for the next 5 epochs. We continue this pattern, unfreezing one additional ResNet block every 5 epochs until epoch 15. We never propagate through the first block as we find this decreases the overall accuracy, likely due to overfitting. We use standard cross entropy loss and a batch size of 256. Training takes $\sim$1 hour on eight 12GB TITAN V GPUs. 

\section{Conditional Situation Recognition}
\label{sec:csl}

\begin{figure}
\centering
\includegraphics[height=5cm]{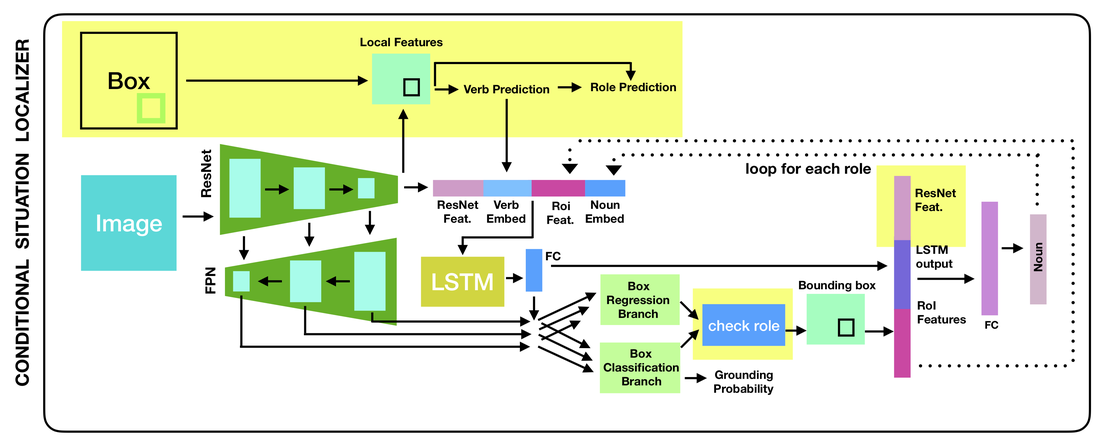}
\caption{Model schematics for the CSL model. Differences between JSL and CSL are highlighted in yellow.
}

\label{fig:csl}
\end{figure}

The Conditional Situation Localizer (CSL) is a modification of JSL which conditions its output on a specific bounding-box, as illustrated by Figure \ref{fig:conditional}. The network architecture of CSL is illustrated by Figure \ref{fig:csl}, with differences from JSL highlighted in yellow. Rather than predicting the verb via a separate network, the verb prediction is done from the local features inside the bounding-box. As in JSL, these local features are obtained by performing RoI Align on the last spatial features of ResNet. Then the verb prediction and these local features are used to predict the role that the object within the box plays with respect to the verb. For example in Figure \ref{fig:conditional}A1, the local features surrounding the query were first used to predict the action as `Calling' and then this verb prediction and those local features where used to predict that the object in the bounding-box fills the second role for this verb, which corresponds to `Tool' in this case. 

CSL then works exactly as JSL except the input bounding-box is used for the predicted role. So if the model predicts that the bounding-box corresponds to the second role for the predicted verb, then on the second pass of the LSTM, the bounding-box prediction made by the classification and regression branches are overwritten by the position of the input bounding-box. This is demonstrated by the ``check role" portion of Figure \ref{fig:csl}. At each pass, the network checks if the current iteration is equal to the role predicted by the input bounding-box. If it is, then that bounding-box is used, otherwise the predicted bounding-box is used. 
\section{Qualitative}
\label{sec:qual}

\begin{figure}[h!]
\centering
\vspace{2em}
\includegraphics[width=\textwidth]{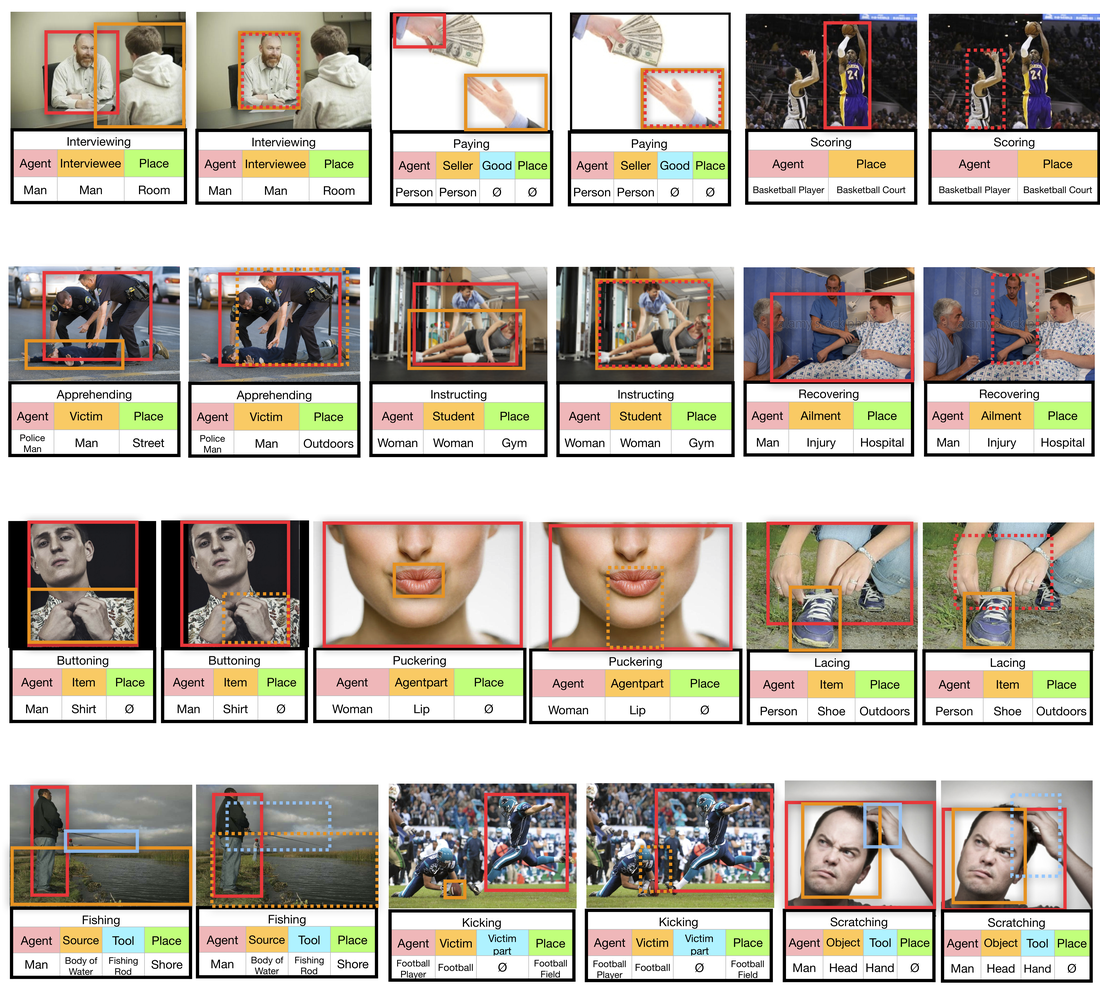}
\caption{For all images, the detections generated by JSL are shown first followed by the detections generated by ISL. Incorrect detections are shown with dotted lines and boxes are colored to correspond with roles. 
}

\label{fig:islvjsl}
\end{figure}

We present additional qualitative results further demonstrating the efficacy of JSL. Figure~\ref{fig:islvjsl} shows a comparison between groundings generated by JSL and ISL for the same image. We illustrate these differences on a sample of images where both ISL and JSL are able to classify the nouns correctly, but ISL fails to correctly locate the entities in the frame. Here we show two common reasons that ISL fails to locate the correct object. The first 2 rows of Figure~\ref{fig:islvjsl} demonstrate the case where there are multiple people in the scene and ISL is unable to pick the correct one for a given role. Because ISL cannot condition its detection on situation, it is often unable to select the correct object when there are multiple objects of the same category present in an image. The bottom 2 rows of Figure~\ref{fig:islvjsl} show cases where ISL correctly locates the object, but fails to create an accurate bounding box around that object. This demonstrates a potential advantage of JSL as predicting the objects in sequence may allow for more accurate localization.

\begin{figure}[h!]
\centering
\includegraphics[width=\textwidth]{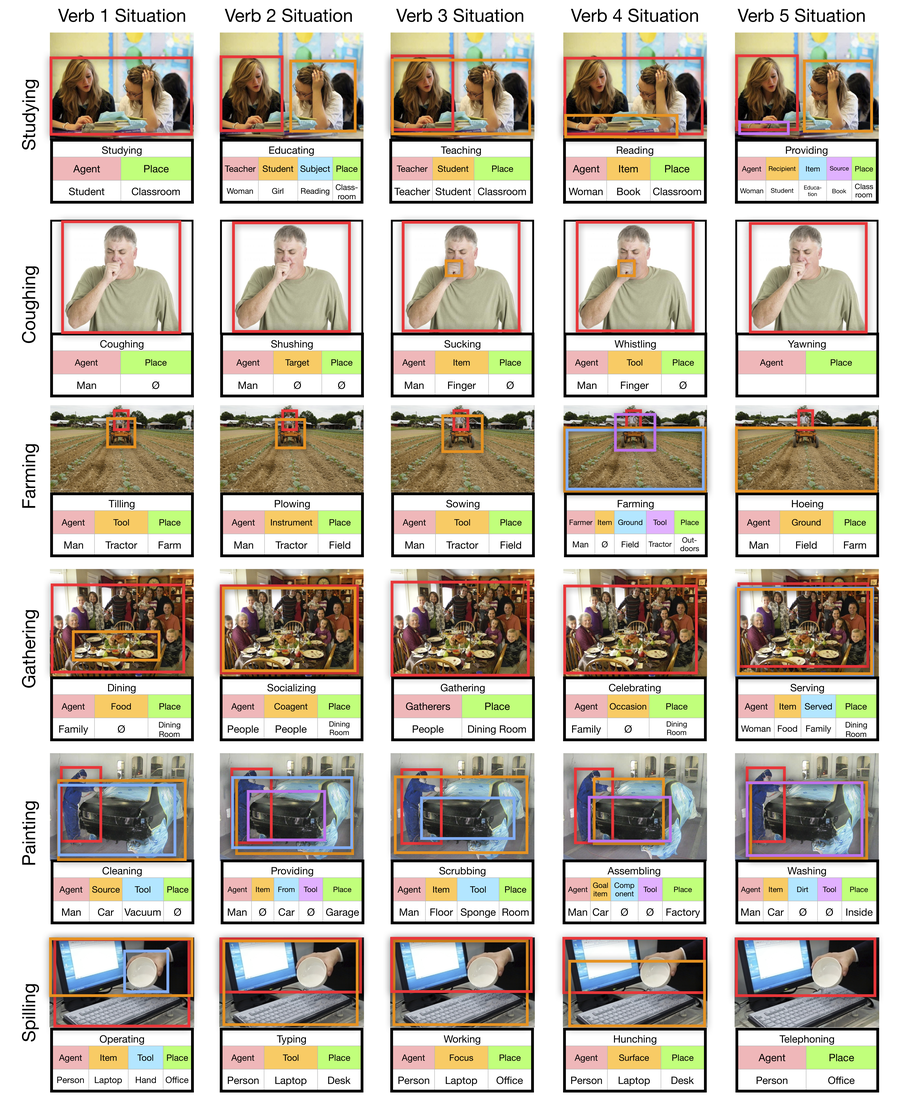}
\caption{Top-5 predictions for a sample of images in the SWiG dev set. The ground truth verb is indicated on the left of each row.}

\label{fig:top5}
\end{figure}

Additionally, Figure~\ref{fig:top5} shows the generated situations for different verbs given the same image. For each image we obtain the top 5 most probable verbs and then generate the grounded situations for each of these verbs. The top two rows of Figure~\ref{fig:top5} are examples where the top verb guess is correct. The first row demonstrates the model's ability to describe the scene in terms of the interaction between two participants as well as what actions they are doing together. In this case, it is clear that both girls are studying, but one is explaining something to the other. Looking at multiple possible verbs captures these complexities. The following two rows are examples where the correct verb is in the top 5 and the bottom two rows show examples where the correct verb is not in the top 5. This tends to happen when the action is very unusual or occurring in a strange context, such as purposefully spilling a cup of water on a keyboard.

\clearpage
\end{document}